\documentclass{article}

\usepackage[final]{corl_2022} %

\usepackage[utf8]{inputenc} %
\usepackage[T1]{fontenc}    %
\usepackage{hyperref}       %
\usepackage{url}            %
\usepackage{booktabs}       %
\usepackage{amsfonts}       %
\usepackage{nicefrac}       %
\usepackage{microtype}      %
\usepackage{xcolor}         %
\usepackage{enumitem}
\usepackage{multirow}
\usepackage{array}
\newcolumntype{P}[1]{>{\centering\arraybackslash}p{#1}}
\usepackage{algorithmic}
\usepackage[algo2e, ruled]{algorithm2e}
\usepackage{algorithm}%
\usepackage{enumitem,kantlipsum}
\usepackage{graphicx,wrapfig,lipsum} 
\usepackage{epsfig} %
\usepackage{mathptmx} %
\usepackage{times} %
\usepackage{amsmath} %
\usepackage{amssymb}  %
\usepackage{algorithmic}
\usepackage[algo2e, ruled]{algorithm2e}
\usepackage{algorithm}%
\usepackage{relsize}
\usepackage[font=small]{caption}
\usepackage{wrapfig}
\usepackage[capitalise]{cleveref}
\usepackage{newtxtext, newtxmath}
\usepackage[export]{adjustbox}
\usepackage{mathtools}
\usepackage{relsize}
\usepackage{authblk}
\renewcommand{\footnotesize}{\fontsize{8pt}{11pt}\selectfont}
\usepackage{hyperref}

\DeclareSymbolFont{rsfs}{U}{rsfs}{m}{n}
\DeclareSymbolFontAlphabet{\mathscrsfs}{rsfs}
\DeclareMathOperator*{\argmax}{arg\,max}

\makeatletter
\renewcommand\AB@affilsepx{, \protect\Affilfont}
\makeatother

\title{Learning Representations that Enable \\ Generalization in Assistive Tasks}

\author[1]{Jerry Zhi-Yang He}
\author[2]{Aditi Raghunathan}
\author[3]{Daniel S. Brown}
\author[2]{\\Zackory Erickson}
\author[1]{Anca D. Dragan}
\affil[1]{University of California Berkeley}
\affil[2]{Carnegie Mellon University}
\affil[3]{University of Utah}
\affil[ ]{\protect\\ \texttt {\{hzyjerry,anca\}@berkeley.edu}, \texttt {dsbrown@cs.utah.edu}, \texttt {\{raditi,zerickso\}@cmu.edu}}

\begin{document}
\maketitle

\newcommand{\RR}{\mathbb{R}}
\newcommand{\Tau}{\mathcal{T}}
\newcommand{\Expect}{\mathbb{E}}
\newcommand{\Loss}{\mathcal{L}}

\definecolor{ao(english)}{rgb}{0.0, 0.5, 0.0}

\newcommand{\pihuman}{\pi_{\textnormal{H}}}
\newcommand{\pirobot}{\pi_{\textnormal{R}}}

\newcommand{\dtrain}{\mathscrsfs{D}_{\textnormal{train}}}
\newcommand{\dtest}{\mathscrsfs{D}_{\textnormal{test}}}
\newcommand{\dist}{\mathscrsfs{D}}

\newcommand{\encoder}{\mathcal{E}}
\newcommand{\decoder}{\mathcal{D}}

\newcommand{\sref}[1]{Sec. \ref{#1}} %
\newcommand{\figref}[1]{Fig. \ref{#1}} %
\newcommand{\appendixref}[1]{Appendix \ref{#1}}
\renewcommand*{\eqref}[1]{Eq. (\ref{#1})} %

\newcommand{\jhnote}[1]{\textcolor{blue}{JH: #1}}
\newcommand{\adnote}[1]{\textcolor{red}{A: #1}}
\newcommand{\ar}[1]{\textcolor{orange}{AR: #1}}
\newcommand{\dbnote}[1]{\textcolor{purple}{DB: #1}}
\newcommand{\zackory}[1]{\textcolor{cyan}{ZE: #1}}

\newcommand{\algname}{Prediction-based Assistive Latent eMbedding\xspace}
\newcommand{\algabbr}{PALM\xspace}

\vspace{-2em}
\begin{abstract}
Recent work in sim2real has successfully enabled robots to act in physical environments by training in simulation with a diverse ``population'' of environments (i.e. domain randomization). In this work, we focus on enabling generalization in \emph{assistive tasks}: tasks in which the robot is acting to assist a user (e.g. helping someone with motor impairments with bathing or with scratching an itch). Such tasks are particularly interesting relative to prior sim2real successes because the environment now contains a \emph{human who is also acting}. This complicates the problem because the diversity of human users (instead of merely physical environment parameters) is more difficult to capture in a population, thus increasing the likelihood of encountering out-of-distribution (OOD) human policies at test time. We advocate that generalization to such OOD policies benefits from (1) learning a good latent representation for human policies that test-time humans can accurately be mapped to, and (2) making that representation adaptable with test-time interaction data, instead of relying on it to perfectly capture the space of human policies based on the simulated population only. We study how to best learn such a representation by evaluating on purposefully constructed OOD test policies. 
We find that sim2real methods that encode environment (or population) parameters and work well in tasks that robots do in isolation, do not work well in \emph{assistance}.  In assistance, it seems crucial to train the representation based on the \emph{history of interaction} directly, because that is what the robot will have access to at test time. Further, training these representations to then \emph{predict human actions} not only gives them better structure, but also enables them to be fine-tuned at test-time, when the robot observes the partner act. \url{https://adaptive-caregiver.github.io}.

\end{abstract}


\section{Introduction}
	
Our ultimate goal is to enable robots to assist people with day to day tasks. In the context of patients with motor impairments, this might mean assistance with scratching an itch, bathing, or dressing \cite{miller1998assistive,brose2010role,erickson2020assistive}. These are tasks in which doing reinforcement learning from scratch in the real world is not feasible, and so sim2real transfer is an appealing avenue of research. Sim2real methods for physical robot tasks in isolation typically work by constructing a diverse "population" of environments and training policies that can work with any member of the population (e.g. a range of parameters of a physics simulator or a range of lighting and textures) \cite{tobin2017domain,tan2018sim,yu2019sim,kumar2021rma,mahler2017dex,wu2019learning,seita_fabrics_2020,hietala2021closing}.

Similarly, population-based (self-play) training has proven successful in zero-sum games against humans \cite{silver2018general,balduzzi2019open,jaderberg2019human}. But unlike tasks the robot does in isolation, assistance requires coordinating with a human who is also acting. And unlike competitive settings, assuming the human to be optimal when they are not, can result in dramatically poor performance \cite{carroll2019utility}. 
Thus, in sim2real for assistance, we have to design a population of potential users and strategies to train with, akin to the physical environment parameters in typical sim2real tasks, rather than the standard population-based training approaches used in competitive settings. But designing a population that is diverse and useful enough to enable generalization to test-time humans, each with their own preferences, strategies, and capabilities, remains very challenging, making it likely that test-time partners might lie outside of the distribution the population was drawn from. Therefore, sim2real methods for assistance will need to be ready to \emph{generalize} to \emph{out-of-distribution} partner policies.

\begin{figure}[t]
  \centering
  \includegraphics[width=\textwidth]{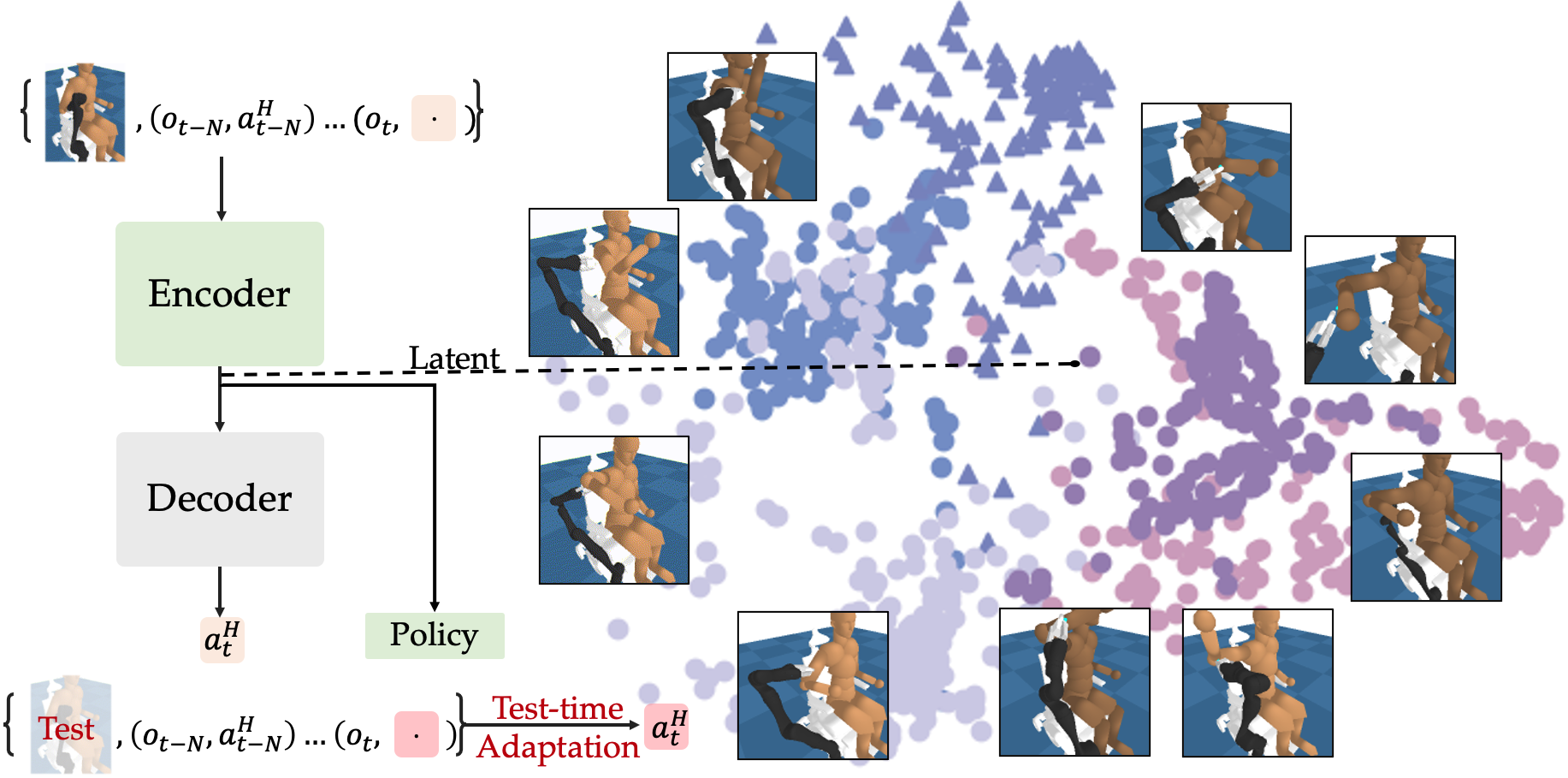}
  	 \small \caption{The framework for jointly learning the Personalized Latent Embedding Space and the robot policy. During training time, we train all components end-to-end to optimize for action prediction (orange) and the robot policy (green). At test time, we can further optimize for this objective to perform test-time adaptation (red). The resulting latent space captures the underlying structure of the preferences and strategies of the training humans.}
	\label{fig:method-architecture:01}
	\vspace{-2em}
\end{figure}

In this work, we identify two principles as key to enabling better generalization. First is that we benefit from learning a latent space of partners that distills their policies down to a structure that is useful for the robot's policy \emph{and} that makes it easy to identify partners at test time. Second is that we need to be prepared for this space to not perfectly capture the space of real human policies, and design it so that it is \emph{adaptable} with real test-time interaction data. 

We thus propose a framework that \emph{learns a latent space directly from history of interaction by predicting the partner's actions}. Our framework allows a robot to capture the relevant information about the human partner that the robot can actually identify when starting to interact, and also enables test-time adaptation of the latent space itself when observing the partner's actions. When evaluated with partner policies we purposefully design to be out-of-distribution, we find that our approach leads to better generalization than prior methods which either do not learn a latent space at all~\cite{strouse2021collaborating}, do not learn a latent space directly based on interaction history~\cite{kumar2021rma}, or train a latent space based on other observables, like states or rewards~\cite{xie2020learning,parekh2022rili}.  Our contributions are four-fold:

\begin{enumerate}[leftmargin=*,topsep=0pt,itemsep=-1ex,partopsep=1ex,parsep=1ex]
    \item We introduce an assistive problem setting where the focus is explicitly on generalization to out-of-distribution partner policies. 
    \item We introduce a framework for training policies for this problem setting, \algname (\algabbr). This enables us to study different methods for learning latent representations on how well they enable generalization.
    \item We identify that the design choice of training a latent space by \emph{predicting partner actions} \emph{directly from history} outperforms (1) state-of-the-art sim2real approaches used in non-assistive tasks that are based on embedding environment parameters \cite{kumar2021rma} as well as (2) human-robot interaction approaches that train representations by predicting observed states or rewards \cite{xie2020learning,parekh2022rili}.
    \item We propose to adapt the learned latent space at test time, upon observing the partner's actions, and show it leads to generalization performance gains.
\end{enumerate}

\section{The Assistive Personalization Problem}
In this section, we introduce the personalization problem in an assistive context. In particular, our goal is to learn a robot policy $\pi_R$ that can assist a novel human partner in zero-shot fashion, or with a small amount of test-time data. 

\textbf{Two-player Dec-POMDP}. An assistive task can be modeled as a two-agent, finite horizon decentralized partially-observable Markov decision process (Dec-POMDP) and is defined by a tuple $\langle S, \alpha,  A_R, A_H, \Tau, \Omega_R, \Omega_H, O, R \rangle$. Here $S$ is the state space and $A_R, A_H$ are the human's and the robot's action spaces, respectively. The human and the robot share a real-valued reward function $R: S \times A_R \times A_H  \rightarrow \RR$; however, we assume that the reward function is not necessarily observed by the robot, i.e. its parameters (e.g. the location of the human has an itch) are in the hidden part of the state. $\Tau: S \times A_R \times A_H \times S \rightarrow [0, 1]$ is the transition function, which outputs the probability of the next state given the current state and all agents' actions. $\Omega_R$ and $\Omega_H$ are the sets of observations for the robot and human, respectively, and $O:S \times A_R \times A_H \rightarrow \Omega_R \times \Omega_H$ represents the observation probabilities. We denote the horizon of the MDP by $T$.

\textbf{Target User.} We target users with partial motor functions --- a common impairment for individuals with partial arm functions. This is an impairment that can occur in some people with cervical SCI, ALS, MS, and some neurodegenerative diseases --- leading to the need for robotic assistance. We model the extent of the impairment as the privileged information in the Dec-POMDP. The robot does not know this a-priori and thus needs to adapt to individual users' capabilities.

\textbf{The Robotic Caregiving Setup.}
We define the observation space for the robot and the human following \cite{erickson2020assistive}: the robot observes its own joint angles, and the human's joint positions in the world coordinate and contact forces; the human observes their joint angles (proprioception) and the end-effector position of the robot. 
When training with simulated humans, the robot gets a reward signal (which depends on privileged information), and has to use that signal to learn to implicitly identify enough about the human to be useful; at test time, the robot does not observe reward signal and must use what it has learned at training time to identify the human's privileged information and be helpful. %

\textbf{Distributions of Humans.} Let function $\pi_H: \Omega_H^* \times A_H \rightarrow [0, 1]$ be the human policy that maps from local histories of observations $\mathbf{o}_t^H = (o^H_1, \ldots, o^H_t)$ over $\Omega_H$ to actions. We define two distributions of human policies $\pi_H \in \dtrain, \dtest$. In the assistive itch scratching, $\dtrain$ can be a set of humans with different itch positions on their arms, which lead to their different movements. We refer to them as \textit{in-distribution humans}. $\dtest$ contains \textit{out-of-distribution} humans whose itch position differ from those in the $\dtrain$.
At training time, the robot has access to $\dtrain$. Thus, it has ground-truth knowledge about the training human's privileged information, such as each human's itch position. At test time, we evaluate the robot policy by sampling humans $\pi_H \sim \dtest$ and directly pairing them with the robot policy. We evaluate the zero-shot and few-shot adaptation performance of the robot policy.

\textbf{Objective.} The main problem we study is how to leverage the training distribution to learn a robot policy $\pi_R:\Omega_R^* \rightarrow A_R$ such that we achieve the best performance on test humans. Concretely, we define the performance of the robot and human as 
\begin{equation}
    J(\pi_R, \pi_H) = \mathbb{E}\left[\sum_{t=0}^T R(s_t,\pi_R(\mathbf{o}^R_t), \pi_H(\mathbf{o}^H_t)) \right],
\end{equation}
Only given access to $\dtrain$, our objective is to find the robot policy $\pi_R = \argmax_\pi J(\pi, \pi_H), \pi_H \sim \dtest$.

\section{Learning Personalized Embeddings for Assistance with \algabbr}

In this section, we present \algname (\algabbr). We introduce the general framework of using a latent space to perform personalization in an assistive context. We then highlight the advantage of action prediction in contrast to prior works. Finally, we describe how we can optimize \algabbr at test time to work with out-of-distribution humans.

\subsection{Learning an Assistive Latent Space}
\label{sec:method-training-palm}
Given a training distribution of humans $\dtrain$ \footnote{we describe how we generate this distribution in \sref{exp:environments-intro}}, we would like to learn a robot policy that can adapt to assist new users. To achieve that, a robot must learn to solve the task while efficiently inferring the hidden component that differs across humans. One natural way to do so is to \emph{learn a latent space} that succinctly captures what differs across humans in a way that affects the robot's policy. When deployed on a test human, the robot infers this latent embedding and uses it to generate personalized assistance.

We denote the latent space as $z_t \sim \encoder_\theta( z; \tau_{1:t})$, where $\encoder_\theta$ encodes the trajectory $\tau$ so far and outputs latent vector $z_t$. The robot uses this latent space to compute its actions $a_t^R = \pi_R(o_t^R, z_t)$. We train the base policy $\pi_R$ and the latent space encoder $\encoder_\theta$ jointly as they are interdependent \cite{liu2021decoupling} --- better robot policy leads to different trajectories across humans, which in turn leads to distinguishing $z$. See \appendixref{appendix_sec:implementation} on training details. Ideally, we would like $z$ to capture sufficient information to differentiate the humans, similar to performing a ``dimensionality reduction'' on human policies $\pi_H$. We hereby introduce different objectives for learning such latent space, and how our method --- learning by action prediction --- makes a good fit for the assistive personalization problem.

\subsection{How to Construct the Latent Space}

\noindent \textbf{Prior work and limitations} LILI~\cite{xie2020learning} and RILI~\cite{parekh2022rili} learn a latent embedding of the humans by predicting the next observations and rewards. While they have been shown to work in predicting and influencing human behaviours, both methods assume access to the reward function at test time, which we do not have access to in the assistive setting --- we don't know a-priori the preference and needs of a new user. RMA ~\cite{kumar2021rma} enables fast robot adaptation by learning a latent space of environment parameters, such as friction, payload, terrain type, etc. While it works for a single robot, it is unclear in human-robot settings, how we can encode human motions and preferences as environment parameters.

\textbf{Learning by action prediction.} Given history $\tau_{t-N:t} = \big( (o^R_{t-N},a^H_{t-N}), \ldots (o^R_t, \cdot) \big)$ of $N$ robot observation and human action\footnote{\footnotesize We do not assume access to the person's sensorimotor action (e.g. joint torques). We define human action as change in the person's Cartesian pose, which can be tracked externally.} 
pairs, as outlined in \figref{fig:method-architecture:01}, we embed this trajectory to a low-dimensional manifold and use it to predict $a^H_t$. The intuition is that if we are able to predict the next action by this human, we extract the sufficient information about the human's policy $\pi_{H}$. The latent vector $z$ is representative of the trajectory so far and indicative of the person's future actions. We do this by training a decoder $\decoder_\phi$ parameterized by $\phi$ to predict the next action from the encoder's output $z \sim \encoder_\theta( z; \tau_{1:t})$. 
\begin{equation}
    \Loss_{\textnormal{pred}} = \min_{\theta, \phi}|| \decoder_\phi(z)  - a^H_{t+1}||^2 + c_{\textnormal{KL}} \cdot KL(\encoder_\theta( z ; \tau_{1:t}) || \mathcal{N}(z))
    \label{eq:encoder-loss}
\end{equation}
The encoder $\encoder$ is a recurrent neural network parameterized by $\theta$. Here the second term is a regularization term motivated by Variational Autoencoder \cite{kingma13vae, watter15embedtocontrol}, that enforces the latent space to follow a normal prior distribution. This encourages nearby terms in the latent space to encode similar semantic meanings. In the context of assistive tasks, this helps us better cluster similar humans closer in the latent space, and we show a didactic example in \sref{sec:didactic-qualitative-toy}.

\subsection{Latent Space Adaptation at Test Time}
\label{method:test-time-opt}
At test time, as we work with a new user, we would like our encoding of the new user to match the true latent information, $z^*$ of that user. In other words, we would like to minimize $|| \encoder(\tau) - z^* ||^2$. Because we do not know about the new users a-priori, we can only optimize for this objective via proxy, which we refer to as \textit{test time adaptation}.

Since the \algabbr latent space is based on action prediction, we can adapt it to a new user by further optimizing the latent space. Note that because \eqref{eq:encoder-loss} requires only observation-action data, we do not need any additional label to perform test time adaptation. More formally, we collect a small dataset of test-time interaction trajectories, $\tau$, and perform a few gradient steps to optimize both the encoder and decoder for \eqref{eq:encoder-loss}: $(\theta, \phi) \rightarrow (\theta, \phi) - \delta \nabla_{(\theta, \phi)} \Loss_{\textnormal{pred}}(\encoder_\theta, \decoder_\phi, \tau)$. 

The idea of test-time optimization has been shown to improve perceptual robustness for grasping in sim2real research \cite{yoneda2022invariance}. We follow a similar pipeline, where we can improve the latent encoding by collecting unsupervised action data from test users. Here the main difference is instead of perceptual differences, our goal is to reduce the domain gap on test users.

\section{Experiments}
\label{sec:experiments}

In this section, we evaluate our method \algabbr (\algname) in collaborative human-robot environments of varying tasks and varying populations of human models. In particular, we focus on the out-of-distribution generalization by constructing different forms of out-of-distribution populations. We focus on empirically investigating the benefits of learning a latent space, the effect of different kinds of prediction on learning a useful latent space, the properties of learned latent spaces, and the gains from test-time adaptation to humans.

\begin{wrapfigure}{r}{7cm}
\vspace{-0.5cm}
\includegraphics[width=7.5cm]{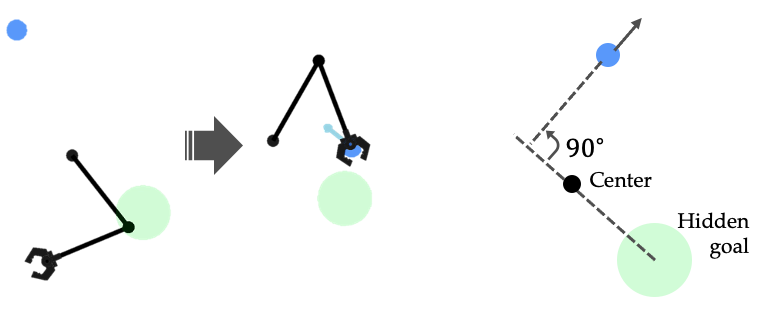}
\small \caption{The assistive reacher environment. Left: the robot's goal is to move the human agent towards the hidden target. Right: the hidden goal's position can be inferred 90 degrees from the humans force output.}
\label{fig:reacherenv}
\vspace{-1mm}
\end{wrapfigure}

\subsection{Environments}
\label{exp:environments-intro}
Here we introduce two environments where we study assistive personalization. In both environments, the robot has to infer some hidden information from the human in order to successfully solve the task. Note that these are examples meant for demonstrating the effectiveness of the algorithm, and we do not claim to solve the full robotic caregiving problem.

\textbf{Assistive Reacher} (\figref{fig:reacherenv}) is 2D collaborative environment where a two-link robot arm assists a point human agent to get to the target position. This target is located at $(d \cos{\alpha_{\textnormal{H}}}, d \sin{\alpha_{\textnormal{H}}})$, where $d$ is a fixed value, and $\alpha_{\textnormal{H}}\in [ -\pi, \pi]$ is known to the human, but not the robot. The human agent is initialized randomly in the 2D plane with random hidden parameters $\alpha_\textnormal{H} \in [-\pi, \pi], k_\textnormal{H} \in [0.5, 1.5]$. The robot can only identify the target position by physical interactions --- once the robot initiates contact, the human applies a force  $ k_\textnormal{H} \cdot (\cos{\alpha_{\textnormal{H}} + \frac{\pi}{2}}, \sin{\alpha_{\textnormal{H}}   + \frac{\pi}{2}})$. Only by recognizing the human in terms of $\alpha_{\textnormal{H}}, k_\textnormal{H}$ can the robot compensate the force, and successfully move the human to the hidden target. Each episode has 40 timesteps.

\textbf{The Scope of Generalization.} We define $\dtrain$ as 36 samples uniformly sampled from $\alpha_\textnormal{H} \in [-\pi, \frac{\pi}{2}], k_\textnormal{H} \in [0.5, 1.5]$ and $\dtest$ as 12 samples uniformly sampled from $\alpha_\textnormal{H} \in [\frac{\pi}{2}, \pi], k_\textnormal{H} \in [0.5, 1.5]$.

\textbf{Assistive Itch Scratching} (\figref{fig:method-architecture:01}) is adapted from assistive gym \cite{erickson2020assistive}. It consists of a human and a wheelchair-mounted 7-dof Jaco robot arm. The human has limited mobility --- they can only move the 10 joints on the right arm and upper chest, and needs the robot's assistance to scratch the itch. An itch spot is randomly generated on the human's right arm. The robot does not directly observe the itch spot, and relies on interaction with the human to infer its location. Each episode has 100 timesteps.

We use co-optimization to create $\dtrain$ and $\dtest$ for Assistive Itch Scratching.
A benefit of the co-optimization framework is that it naturally induces reward-seeking behaviour from the human and the robot, which simulates assistance scenarios. For instance, to generate more inactive human policies, we can introduce a weighting term in the reward function for human action penalties $R_\textnormal{p} = c_\textnormal{p} \cdot || \pihuman (s_t)||^2$ where $c_\textnormal{p}$ is a constant controlling the penalty. The overall objective becomes
\begin{align}
\max_{\pihuman, \pirobot} \Expect \big[ \sum_t R \big( s_t, \pihuman (s_t),\pirobot(s_t) \big) \big] + c_\textnormal{p} \cdot || \pihuman (s_t)||^2
\end{align}
\begin{wrapfigure}{r}{3.5cm}
\vspace{-3mm}
\includegraphics[width=3.5cm]{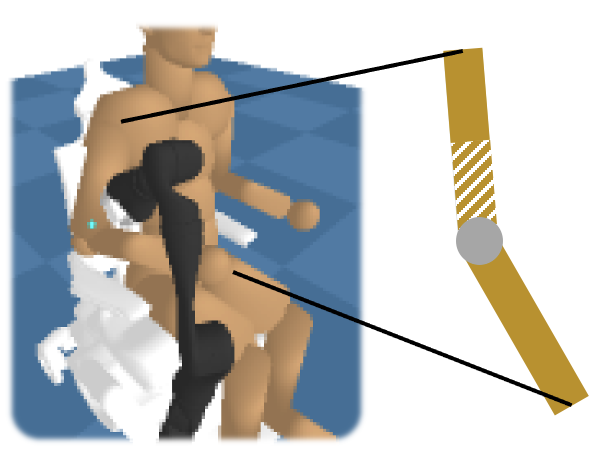}
\small \caption{Definition of $\dtrain$ and $\dtest$.}
\label{fig:exp-arm-definition:1}
\vspace{-3mm}
\end{wrapfigure} 
\textbf{The Scope of Generalization.} We are motivated by real world applications where users tend to have different levels of mobility limitations, or itch locations in different body parts. To generate a synthetic population to capture such diversity in itch scratching task, we explore different co-optimization settings (1) we assign different human action penalty to be $c_{\textnormal{p}} = 3, 3.5, 4$, where larger penalties lead to the human agent exerting less effort. 
(2) We simulate different itch positions on the human's arm and train co-optimized human and robot policies conditioned on them. This leads to qualitatively different strategies for the human and the robot. Note that this serves first step to understanding how different methods generalize, since we never expect to be able to capture the diversity in humans perfectly. 
For training, we use Proximal Policy Optimization (PPO) to optimize human and robot policies in an interleaving fashion. 
Note that we also keep the co-optimized robot policy and use it to obtain expert actions for assistive policy training (see \sref{sec:method-training-palm} and supplement for full details).

To construct $\dtrain$ and $\dtest$, we divide the two arms' areas into four equal portions, as shown in \figref{fig:exp-arm-definition:1}, and generate human policies conditioned on itch positions in these areas. $\dtrain$ consists of three of the four portions and $\dtest$ consists of the remaining one. We then construct three distribution sets in increasing order of difficulties. In the first distribution $\dist^1$, we confine itch positions from a line-shaped region. In $\dist^2$, we sample from all the arm areas. Note that $\dist^1$ and $\dist^2$ are constructed by setting action penalty $c_{\textnormal{p}} = 3$.  In $\dist^3$, we combine humans of $c_{\textnormal{p}} = 3, 3.5, 4$, each trained with two different random seeds. This adds the extra complexity of human activity levels. We simulate 12 in-distribution humans from each of the three training portions under each action penalty. 

\subsection{Baselines}
\label{sec:baselines}
We compare with baselines that do not learn an explicit latent space as well as existing methods for adaptation via learning latent embeddings.

\textbf{MLP and RNN.} We follow \cite{strouse2021collaborating} that trains sequential models to enable adaptation to simulated humans. We explore using a recurrent neural network or a feed-forward network on concatenated state-action histories. The models directly output robot action, and there is no latent space modeling.

\begin{wrapfigure}{r}{5.5cm}
\includegraphics[width=6cm]{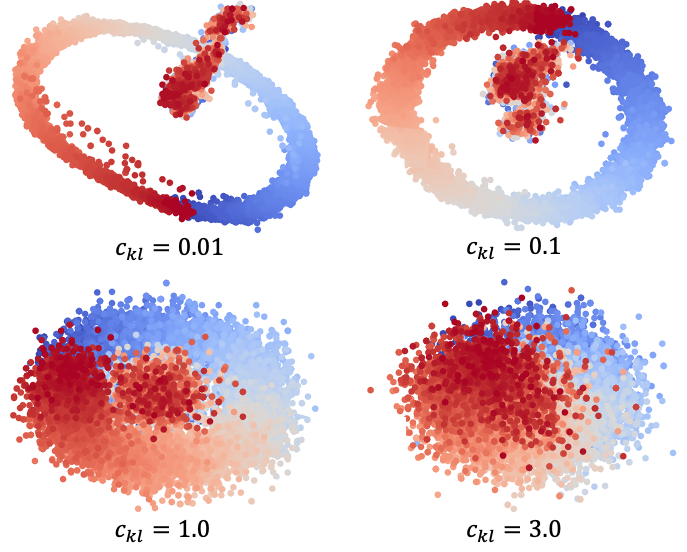}
\small \caption{Latent space of \algabbr in the assistive reacher environment when we can sample humans $\alpha_{\textnormal{H}} \in [ -\pi\text{ (red)}, \pi\text{ (blue)}]$ continuously.}
\label{fig:results-qualitative-human:1}
\vspace{-4mm}
\end{wrapfigure} 

\textbf{ID-based Human Embeddings.} In contrast to learning latent space from history, another class of method studies encodes human-designed environment parameters \cite{kumar2021rma, liu2021decoupling, Kalashnikov21MTOPT}. We focus specifically on RMA \cite{kumar2021rma}, a two-phased method that first learns to encode task-ID (phase I) and then trains a recurrent network to regress to the embeddings from observation history (phase II). For training a quadruped robot, RMA encodes the physical parameters (friction, payload, etc) of the environment. The first stage trains a policy with ground-truth information, and the second phase performs environment identification. While RMA is shown to be effective for learning policy for in-distribution environments, it is unclear how well it generalizes to out-of-distribution environments. Furthermore, in assistive tasks, it is unclear how to construct the "ground-truth ID" for phase I that quantifies the user characteristics. We study \textbf{RMA-Param} and \textbf{RMA-Onehot}, where we assign each training human a one-hot vector. For RMA-Param, we use a three-dimensional vector that includes the $x, y$ position of the itch position and the arm index. When there are multiple human activity levels as mentioned in \sref{exp:environments-intro}, we introduce a fourth dimension with an integer to indicate the action penalty $c_\textnormal{p}$. 

\textbf{LILI and RILI.} We consider two other methods of the \algabbr framework: LILI \cite{xie2020learning} and RILI \cite{parekh2022rili}. As mentioned in \sref{sec:method-training-palm}, LILI jointly predicts future observation and reward, and RILI predicts reward. Given that reward is only available at training time, we cannot perform test time optimization for LILI and RILI.

\textbf{Ablations of \algabbr.} Our method has several components: we use a recurrent neural network to encode interaction history, and use its output to minimize prediction loss $\Loss_{\textnormal{pred}}$ and policy loss $\Loss_{\textnormal{pol}}$. We also use the KL term in \eqref{eq:encoder-loss} to regularize the human embeddings. To test the effectiveness of our method, we separate different parts and create a set of baselines. We hereby describe them in detail: (1) No $\Loss_{\textnormal{pred}}$: the model shares the same encoder and policy network architecture, yet we don't optimize for $\Loss_{\textnormal{pred}}$. By removing the prediction loss in \figref{fig:method-architecture:01}, the latent space is not explicitly trained to contain human information. (2) No $c_\textnormal{KL}$: no regularization in the latent space. (3) Frozen embedding: instead of jointly training embeddings and the policy network, we first train the encoder on expert data, freeze it, and then train robot policy. We include an ablation study of our main experiments in the appendix.

\subsection{Didactic Experiment in Assistive Reacher Environment}
\label{sec:didactic-qualitative-toy}

\textbf{Can \algabbr learn a meaningful distribution from the interaction?} Unlike other ID-based methods like RMA, \algabbr does not have access to the human parameters at training time. We study whether \algabbr can learn a meaningful latent space without explicitly knowing this information. We sample training humans from $\alpha_{\textnormal{H}} \in [ -\pi, \pi]$ continuously. We train \algabbr with different amount of prior regularization, $c_{KL}$ from \eqref{eq:encoder-loss}. We train using a recurrent window of length 4 and a batch size of 512 episodes. Additional training details can be found in the supplementary material.

We average test results using 100 episodes and visualize the results in \figref{sec:didactic-qualitative-toy}. Given that humans are parameterized by $\alpha_\textnormal{H}, K_{H}$, the ideal embedding space looks like a ring with a small blob in the center. The ring corresponds to the 2D projection of $\alpha_{\textnormal{H}}$ and the blob denotes the initial part of interaction before contact, which is indistinguishable. We find that while \algabbr never observes the underlying parameter $\alpha_{\textnormal{H}}$, it can learn a latent space that characterizes $\alpha_{\textnormal{H}}$. Interestingly, varying the amount of regularization qualitatively affects the shape of the latent space. Setting the VAE regularization $c_{\textnormal{KL}} = 0.1$ recovers a latent space that most resembles to the ideal latent space.

\subsection{Assistive Reacher Main Experiment}

\noindent\textbf{Experiment Setting.} We use the finite $\dtrain$ described in \sref{exp:environments-intro} and train all baselines for 200 epochs with 512 batch size. We then evaluate the trained policies on $\dtest$. We normalize the resulting reward with respect to oracle reward.

\noindent\textbf{Results.} We average test results using 100 episodes. On in-distribution humans, we find that all methods successfully follow the right policy that assists the human to reach their goal. This shows that they all successfully predict the human latent information explicitly or implicitly. On out-of-distribution humans, the methods are no longer guaranteed to predict the correct embedding. \algabbr with action prediction significantly outperforms other methods. With test-time adaptation, \algabbr further improves. 

\begin{figure}[t]
  \centering
  \includegraphics[width=\textwidth]{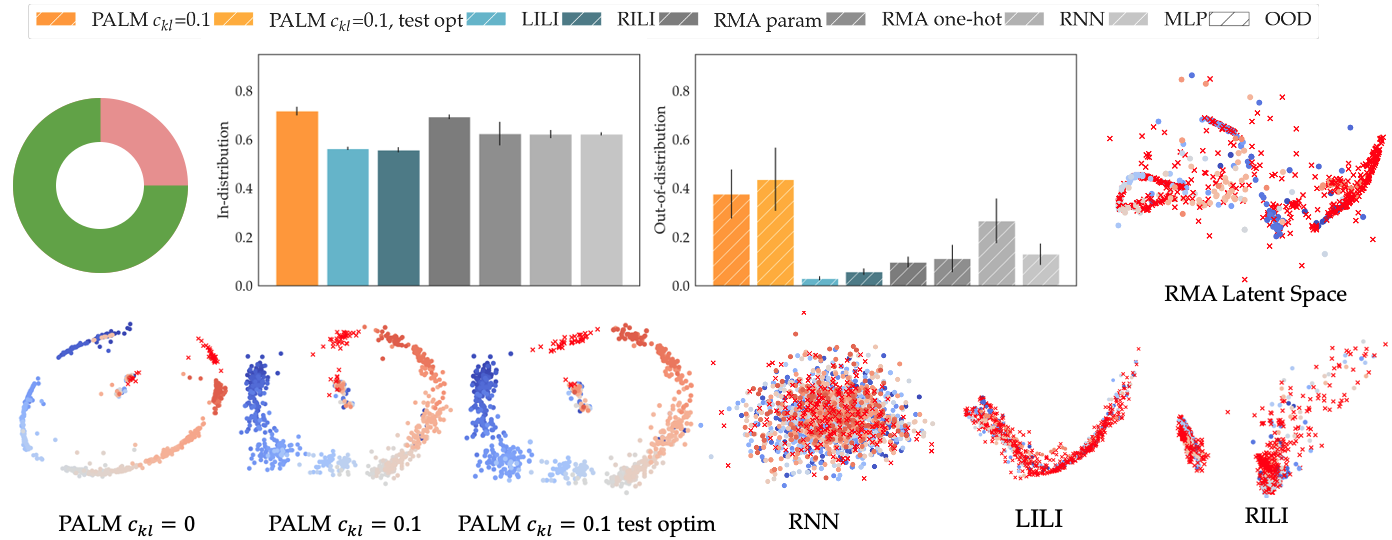}
  	 \small \caption{Top left: evaluation of \algabbr and baselines on in-distribution (green) and out-of-distribution (pink) humans. Right and bottom: visualization of the latent embeddings of different methods. OOD humans are highlighted in red crosses. Best viewed electronically.
  	 }
	\label{fig:exp-toy-result:1}
	\vspace{-0.6cm}
\end{figure}

\noindent\textbf{Visualizing the latent space.} We qualitatively study generalization by visualizing the latent space as well as the mapped embeddings of both in-distribution and out-of-distribution humans (in red crosses) in \figref{fig:exp-toy-result:1}. Interestingly, only \algabbr with action prediction can infer the ``ring'' structure. RMA, RNN, LILI and RILI fail to do so. We hypothesize that because hand-crafted human IDs do not convey the information about human policy, RMA warped the IDs in arbitrary what that are harmful for generalization. The same happens with RILI and LILI. We hypothesize this is due to the inherent ambiguity in reward prediction: a low reward does not necessarily recover the human policy structure.

The visualization also offers some insight into why having $c_{\textnormal{KL}}$ regularization is helpful for generalization. Compare the latent space of \algabbr $c_{\textnormal{KL}}=0$ and \algabbr $c_{\textnormal{KL}}=0.1$, the latter induces a smoother distribution where test humans are better fitting in the ``missing arc'' of the ``ring". Further more, we see that with test time optimization, the \algabbr latent space embeds the OOD human better, by filling in more of the arc.

\subsection{Assistive Itch Scratch Main Experiment}

\begin{figure}[t]
  \centering
  \includegraphics[width=\textwidth]{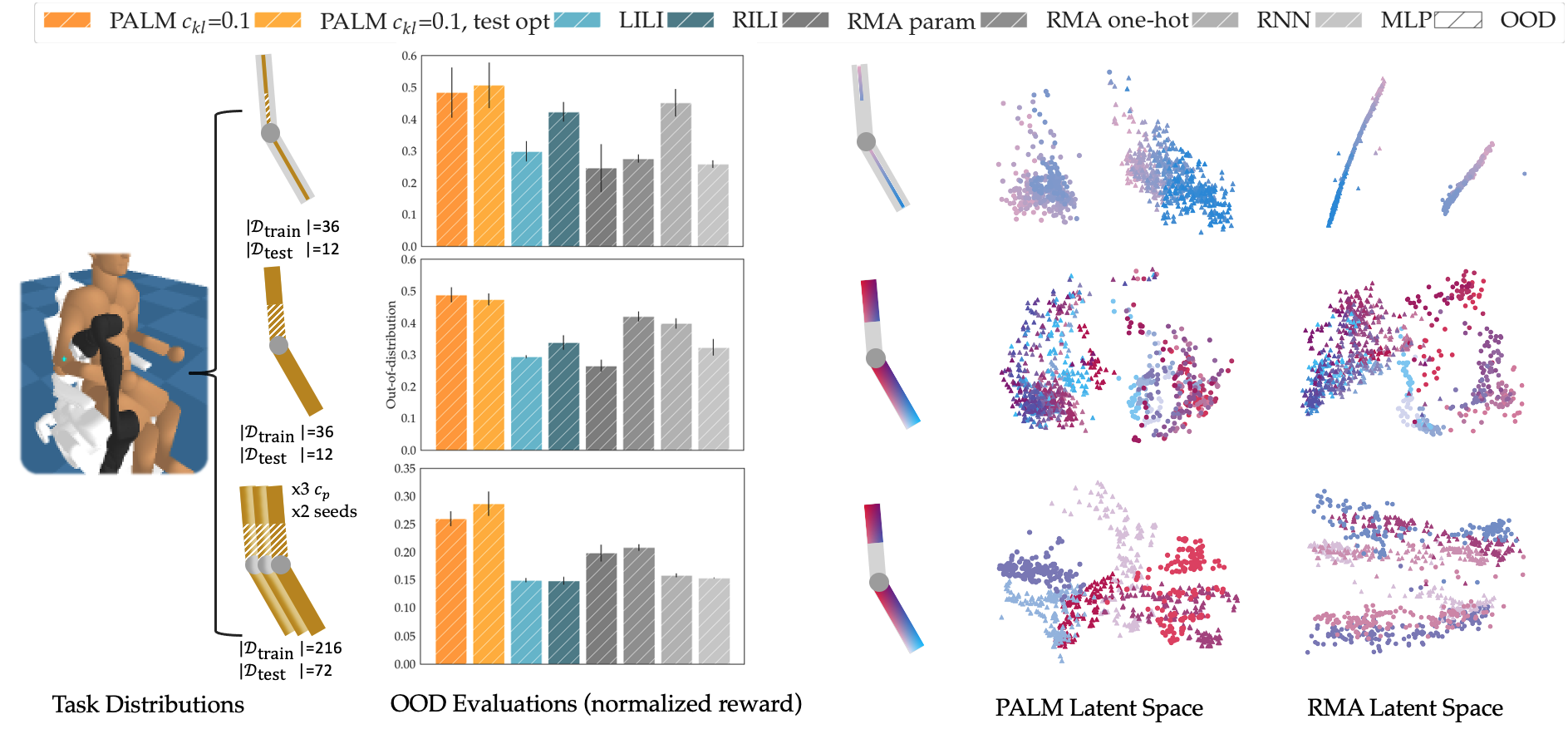}
  	 \small \caption{In assistive itch scratching, we sample humans by different itch positions and activity levels (varying action penalty $c_{\textnormal{p}}$). We visualize the in- and out-of-distribution humans on the two-link arm figures. We also visualize the embedding space of \algabbr and RMA, where we color-code the embeddings of in-distribution humans. Here we leave the embeddings of other baselines to supplementary material.}
	\label{fig:exp-assist-result}%
	\vspace{-0.6cm}
\end{figure}

\noindent\textbf{Results.} We follow a similar procedure as the reach environment to train itch scratching policy, and train for 240 epochs with 192 trajectories for batch size. As shown in \figref{fig:exp-assist-result}, we observe \algabbr with action prediction has better generalization performance than other baselines. We see that in the simplified distribution $\dist^1$, RILI and MLP have the best generalization performance among baselines, yet as the complexity of the training human distribution increases, they deteriorate. Detailed results of the ablation study are included in the appendix.

\begin{wrapfigure}{r}{6.5cm}
\vspace{-3mm}
\includegraphics[width=6.5cm]{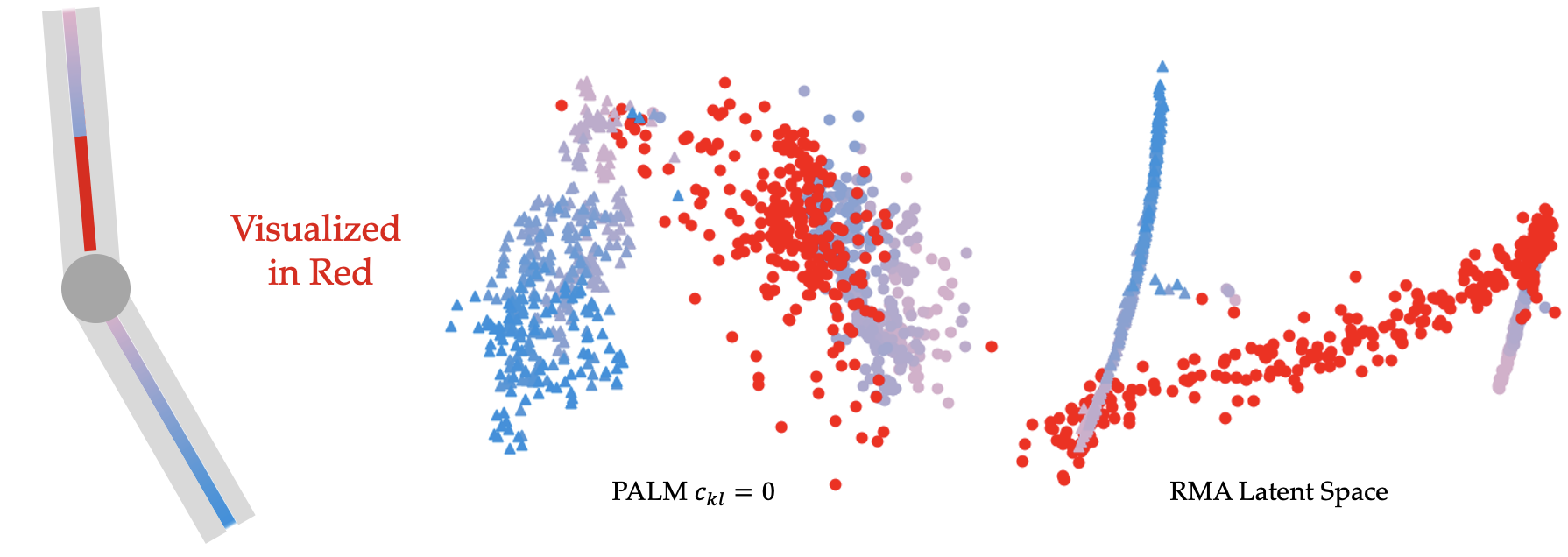}
\small \caption{To better under extrapolations to OOD human, which have new itch locations highlighted in red on the left, we visualize the embeddings of both the IND (colored) and OOD (red) humans on the right. }
\label{fig:eval-latent:1}
\vspace{-3mm}
\end{wrapfigure} 

\textbf{Visualizing the Latent Space} To further investigate why PALM generalize better to OOD human than RMA baselines, We visualize the latent space of the "straight-line" distribution. As we see in ~\figref{fig:eval-latent:1}, \algabbr can capture the structure in human training distribution as two clusters, and also correctly embed the OOD humans distribution as a part of the upper arm distribution. RMA-based methods, on the other hand, can discover the structure of training humans. Yet qualitatively, they fail the correctly embed the OOD humans in proximity to the upper arm distribution.

\vspace{-3mm}
\subsection{Limitations and Failure Cases}
Although \algabbr achieves good average-case performance, it works best with humans sampled near the training distribution. If we pair the robot with an adversarial human, \algabbr is likely to fail as it lacks a fall-back safety policy.

The major limitation of \algabbr is the requirement of generating a human population. While we provide one way to generate human populations based on weighted human-robot co-optimization, we lack ways to systematically generate diverse and realistic human motions. One important direction for future work is to incorporate real user data to create training populations. Improving the realism of the training human population is likely a crucial step to supporting transfer to real partners.

One future direction is extending to settings with one patient and one human caregiver. While our framework still applies, this leads to new challenges including (1) learning a joint or separate latent space for human patient and caregiver, (2) modeling a population of human caregivers for training in simulation, and (3) modeling communication between the human caregiver and patient.

\section{Conclusion}
\label{sec:conclusion}

Generalization is an important task for assistive robotics, and in this paper, we formulate a problem setting that focuses on Out-Of-Distribution users. To that end, we contribute a framework \algabbr for learning a robot policy that can quickly adapt to new partners at test time. \algabbr assumes a distribution of training humans and constructs an embedding space for them by learning to predict partner actions. We can further adapt this embedding at test time for new partners. Experiments show that \algabbr outperforms state-of-the-art approaches. We are excited by the potential of using \algabbr to enable robotic assistance in the future.

\clearpage
\acknowledgments{We would like to thank Ashish Kumar for discussions on the RMA method and Annie Xie for providing the implementation for LILI. We would also like to thank Charlie C. Kemp for feedbacks and insights on assistive tasks. This research was supported by the NSF National Robotics Initiative. }

\bibliography{main}  %

\appendix
\pagenumbering{arabic}

\begin{center}
\textbf{\large Supplementary Material}
\end{center}

In this supplementary material, We first present details and visualizations on how we generated $\dtrain$, the training distribution of human agents in assistive itch scratching in \sref{appendix_sec:population}. We then present the main algorithm pseudo-code for PALM in \sref{appendix_sec:algorithm}. Next we provide implementation details of different methods in our main experiments in \sref{appendix_sec:implementation} and the ablation study of PALM in \sref{appendix_sec:ablation} and an evaluation of PALM with a second task, bed bathing assistance, in \sref{appendix_sec:additional_bed}.

\section{Generating Human Populations}
\label{appendix_sec:population}
To train our robot using sim2real, we would like to have a set of diverse environments. However, unlike single-agent domain randomization where we can vary environment parameters such as friction, in assistive tasks the environment entails a changing user policy. 
It is not obvious how to best generate a diverse population that captures user preferences, levels of disabilities, or movement characteristics. Generating human motions that realistically capture the variation observed in physical human-robot interaction has remained an unsolved challenge in robotics.

\begin{figure}[h]
  \centering
    \includegraphics[width=\linewidth]{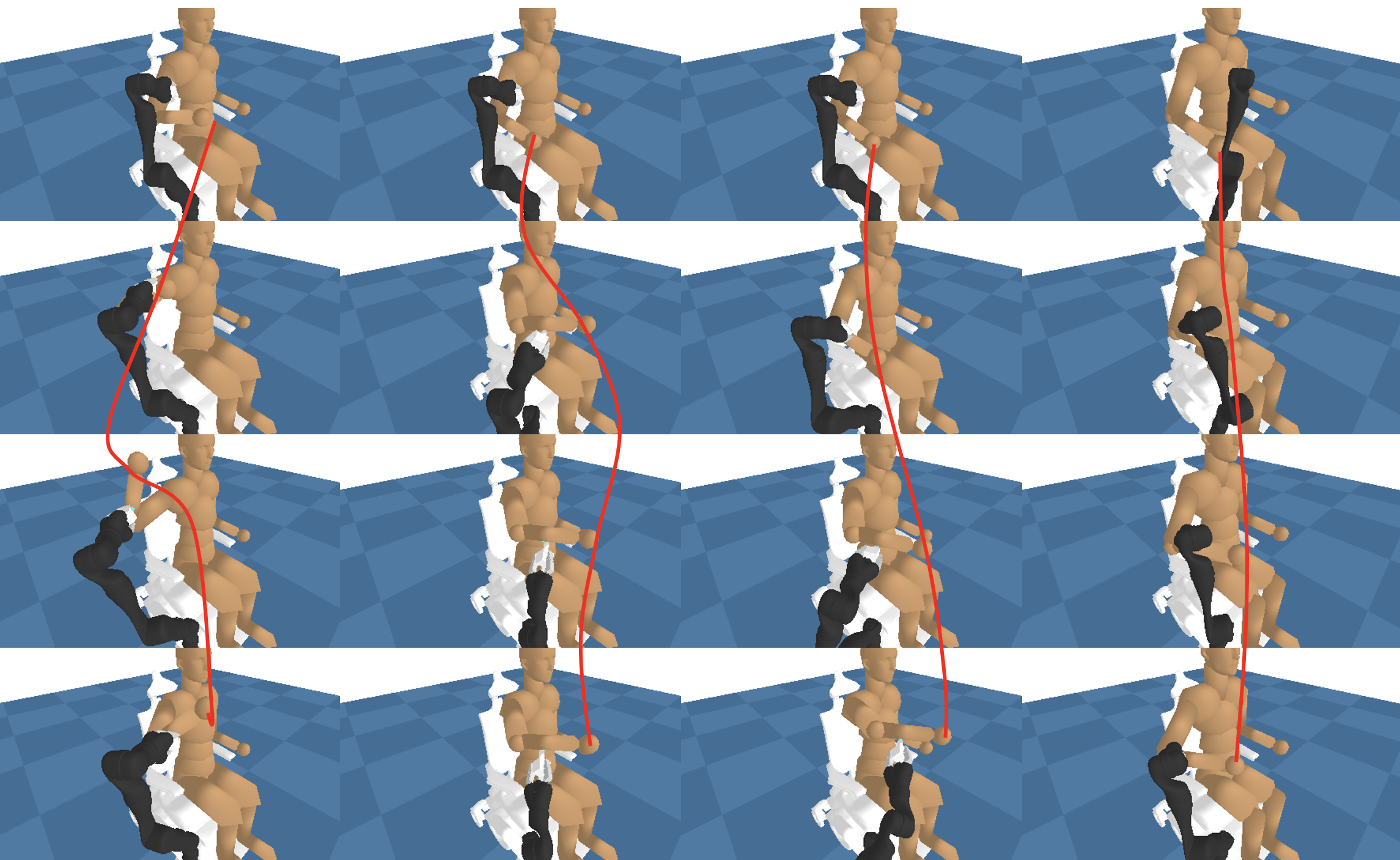}
    \small \caption{Visualization of humans generated of different activity levels. From left to right we apply action penalties $c_{\textnormal{p}} = 0, 10, 30, 60$. Qualitatively, increasing the penalty results in the human taking more steady actions with less swinging motions. This results in the human being more likely to expose the itch spot for the robot to scratch, as opposed to scratching themselves.}
    \label{fig:assistive_itch_humans}
\end{figure}

\textbf{Co-Optimization} Prior works in robotic assistance~\cite{erickson2020assistive, clegg2020learning} have demonstrated that by optimizing for the same task objective, we can generate human and robot motions that coordinate towards the same goal, such as robot-assisted dressing. Further more, we can leverage reward engineering \cite{gupta2022unpacking, zhai2022computational} to generate a diverse set of motions.

To generate diverse population in itch scratching task, we explore two sources of diversities: (1) we assign different human action penalty $c_{\textnormal{p}}$, where larger penalties lead to the human agent exerting less effort. In the simulation experiment we use $c_{\textnormal{p}} = 3, 3.5, 4$. %

(2) We simulate different itch positions on the human's arm and train co-optimized human and robot policies conditioned on them. This leads to qualitatively different strategies for the human and the robot. Note that this serves as the first step to understanding how different methods generalize since we never expect to be able to capture the diversity in humans perfectly. 
For training, we use Proximal Policy Optimization (PPO) to optimize human and robot policies in an interleaving fashion. 
Note that we also keep the co-optimized robot policy and use it to obtain expert actions for assistive policy training (see \sref{sec:method-training-palm} and supplement for full details).

\textbf{Visualization} Here we visualize trajectories from humans with different action penalties in \figref{fig:assistive_itch_humans}. Note that higher penalties result in the human taking more steady actions of smaller magnitude.

\section{Comparison with other VAE baselines for sequential data}

Note that our method relies on embedding human trajectories as sequential data into a latent space. Our implementation uses the final hidden state of RNN as the input to variational autoencoder. This is based on \cite{Bowman15RNNVAE}, which has been shown to be effective in embedding and generating sentences. Given that there are other different generative models for sequential data, our framework can be easily combined with them. In fact, wee believe coming up with a better model for human embedding is a future direction.

We hereby provide comparison with another different generative sequential model \cite{Chung15RNNVAE}. Different from \cite{Bowman15RNNVAE} that uses only the final hidden state, they construct a latent space for every intermediate step in the sequential model. We keep all the experiment hyper-parameters the same, and concatenate the final hidden state with observation as input to the robot policy. We show the results in 

\begin{table}[h]
  \label{table:baseline-rnn-vae}
  \centering
    \begin{tabular}{c|c|c|c|c}
    \hline
      Normalized Reward & Distribution & Our Method \cite{Bowman15RNNVAE} & Baseline \cite{Chung15RNNVAE} & RNN \\ \hline
      Assistive Reacher & 
        \begin{tabular}{@{}c@{}} IND \\ OOD\end{tabular} &  
        \begin{tabular}{@{}c@{}} \textbf{0.72 $\pm$ 0.02} \\ \textbf{0.38 $\pm$ 0.10}\end{tabular} &  
        \begin{tabular}{@{}c@{}} 0.66 $\pm$ 0.02 \\ 0.11 $\pm$ 0.01 \end{tabular} & 
        \begin{tabular}{@{}c@{}} 0.62 $\pm$ 0.02 \\ 0.27 $\pm$ 0.09 \end{tabular} \\\hline
      Itch Scratching $\dist^1$ & 
        \begin{tabular}{@{}c@{}} IND \\ OOD\end{tabular} & 
        \begin{tabular}{@{}c@{}} 0.75 $\pm$ 0.01 \\ \textbf{0.48 $\pm$ 0.08} \end{tabular} &  
        \begin{tabular}{@{}c@{}} 0.45 $\pm$ 0.04 \\ 0.32 $\pm$ 0.01 \end{tabular} &  
        \begin{tabular}{@{}c@{}} \textbf{0.79 $\pm$ 0.01} \\ 0.25 $\pm$ 0.08 \end{tabular} \\\hline
      Itch Scratching $\dist^2$ & 
        \begin{tabular}{@{}c@{}} IND \\ OOD\end{tabular} & 
        \begin{tabular}{@{}c@{}} 0.58 $\pm$ 0.03 \\ \textbf{0.49 $\pm$.0.02} \end{tabular} &  
        \begin{tabular}{@{}c@{}} \textbf{0.70 $\pm$ 0.03} \\ 0.31 $\pm$ 0.05 \end{tabular} &  
        \begin{tabular}{@{}c@{}} 0.44 $\pm$ 0.01 \\ 0.40 $\pm$ 0.02 \end{tabular} \\\hline
      Itch Scratching $\dist^3$ & 
        \begin{tabular}{@{}c@{}} IND \\ OOD\end{tabular} & 
        \begin{tabular}{@{}c@{}} \textbf{0.34 $\pm$ 0.02} \\ \textbf{0.25 $\pm$ 0.01} \end{tabular} &  
        \begin{tabular}{@{}c@{}} 0.23 $\pm$ 0.22 \\ 0.13 $\pm$ 0.01 \end{tabular} &  
        \begin{tabular}{@{}c@{}} 0.18 $\pm$ 0.01 \\ 0.16 $\pm$ 0.01 \end{tabular} \\\hline
    \end{tabular}
    \caption{Comparison with RNN-VAE baseline \cite{Chung15RNNVAE}}
\end{table}

\section{Algorithm}
\label{appendix_sec:algorithm}

We present the main algorithm for PALM assume we have access to training and test distributions $\dtrain$ and $\dtest$.

  \begin{algorithm}[h]
 	\caption{Prediction-based Assistive Latent eMbedding Training}
    \begin{algorithmic}
	
	\STATE Randomly initialize base policy $\pi$, encoder $\encoder_\phi$ parameterized by $\phi$, decoder $\decoder_\theta$ parameterized by $\theta$. Empty replay buffer $D_1$, window size $w$\\
	
	\For{$\textnormal{itr} = 1, .., N_{\textnormal{itr}}$}{
    
    	\For{$i = 1, .., N_{\textnormal{batch}}$}{
    	    Sample $\pi_{\textnormal{H}_i}$ $\sim \dtrain$, optionally find pre-trained expert robot policy $\pi_{\textnormal{E}i}$
    	    
    	    $o_0 \leftarrow \textnormal{env.reset()}$
    	    
    	    Initialize history $H \leftarrow \phi$
    	    
    	    \For{$t = 0, ..., T$} { 
    	    
    	        Get latest $w$ steps from $H$: $H_{-w} \leftarrow H[-w:]$.
    	       
    	        $z_t \leftarrow \encoder_\phi (o_t, H_{-w})$ 
    	        
                $a_{\textnormal{H}t} \leftarrow \pi_{\textnormal{H}_i}(o_0)$
                
                $a_{\textnormal{R}t} \leftarrow \pi(o_t, z_t), a_{\textnormal{e}t} \leftarrow \pi_{\textnormal{E}i}(o_t)$
                
                $o_{t + 1} \leftarrow \textnormal{env.step(} a_{\textnormal{H}t}, a_{\textnormal{R}t} \textnormal{)} $
                
                Store $(o_t, a_{\textnormal{H}t}, a_{\textnormal{R}t}, a_{\textnormal{e}t}, H_{-w})$ in $D_1$
                
    	    }
     	}
     	
     	\For{$j = 1, ..., N_{\textnormal{opt}}$ }{
     	    Sample a batch of $(o_t, a_{\textnormal{H}t}, a_{\textnormal{R}t}, H_{-\tau})$ from $D_1$
     	    
     	    Compute $\Loss_{\textnormal{pred}}$ using \eqref{eq:encoder-loss}, $\Loss_{\textnormal{PPO}}$ using \cite{schulman2017proximal} and $\Loss_{\textnormal{BC}} = \sum_t || a^\textnormal{exp}_t - a^R_t||^2$
     	   
     	    Optimize $\theta, \phi, \pi$ for $\Loss_{\rm\algabbr} = \lambda_{\textnormal{pred}} \Loss_{\textnormal{pred}} + \lambda_{\textnormal{PPO}}\Loss_{\textnormal{PPO}} + \lambda_{\textnormal{BC}}\Loss_{\textnormal{BC}}$
     	    
     	}
 	}
    \end{algorithmic}
  \end{algorithm}

  \begin{algorithm}[h]
 	\caption{Prediction-based Assistive Latent eMbedding Test Time Adaptation}
    \begin{algorithmic}
    
	\STATE Sample $\pi_{\textnormal{H}}$ $\sim \dtest$, Initialize history $H \leftarrow \phi$, Empty trajectory data $\tau$
    
	\For{$i = 0, .., N_{\textnormal{adapt}}$}{
    
    	$o_0 \leftarrow \textnormal{env.reset()}$

        \For{$t = 0, ..., T$}{
        
            Get latest $w$ steps from $H$: $H_{-w} \leftarrow H[-w:]$.
    	       
            $z_t \leftarrow \encoder_\phi (o_t, H_{-w})$ 
            
            $a_{\textnormal{R}t} \leftarrow \pi(o_t, z_t)$
            
            $o_{t + 1} \leftarrow \textnormal{env.step(} a_{\textnormal{H}t}, a_{\textnormal{R}t} \textnormal{)} $
            
            Store $(o_t, a_{\textnormal{H}t}, H_{-w})$ in $\tau$
        
        }
    
        Compute $\Loss_{\textnormal{pred}}$ using \eqref{eq:encoder-loss} on $\tau$, $\theta \rightarrow \theta - \delta \nabla_\theta \Loss_{\textnormal{pred}}(\encoder_\theta, \tau)$.

    }
    \end{algorithmic}
  \end{algorithm}

\section{Additional Training and Implementation Details}
\label{appendix_sec:implementation}

In our experiments in both the assistive reaching and assistive itch scratching, we use a recurrent network over a sliding window of 4-time steps, each of which is a concatenated vector of observation $o_t$, human action $a_{\textnormal{H}t-i}$ and robot actions $a_{\textnormal{R}t-i}$. For the current time step $t$, we use zero vector for $a_{\textnormal{H}t}$ and $a_{\textnormal{R}t}$. We set the latent space dimension to be four, and use a recurrent network with six layers. Our base policy network has four dimensions and hidden size of 100.

\subsection{PPO and Behaviour Cloning} 
We need to train the base policy $\pi_R$ and the latent space encoder $\encoder_\theta( z; \tau)$ jointly because they are interdependent --- $z$ is the input to $\pi_R$, and $\pi_R$ decides the data distribution which leads to $z$.
We simultaneously optimize the prediction loss in \eqref{eq:encoder-loss} and the policy loss using PPO \cite{schulman2017proximal} algorithm. See \appendixref{appendix_sec:implementation} for more training details. To amend for the instability of training with a population of humans, we leverage Behaviour Cloning, where we use expert robot policies obtained via co-optimization (see \appendixref{appendix_sec:population}) to supervise $\pi_R$ on on-policy data.
More specifically, we query the expert actions $a^\textnormal{exp}_t$ in a DAgger fashion\cite{ross2011reduction} \footnote{This ensures that we encounter no distribution shift at deployment time.} during training, and optimize $a^R_t$ to minimize deviation from it: $\Loss_{\textnormal{BC}} = \sum_t || a^\textnormal{exp}_t - a^R_t||^2$. The overall policy optimization loss include three terms: latent prediction loss, PPO loss and the Behaviour Cloning loss: $\Loss_{\rm\algabbr} = \Loss_{\textnormal{pred}} + \Loss_{\textnormal{PPO}} + \Loss_{\textnormal{BC}}$.

\subsection{Hyperparameters} 
\noindent\textbf{PALM Training}  We use $\lambda_{\textnormal{PPO}} = 0.1, \lambda_{\textnormal{BC}} = 1, \lambda_{\textnormal{pred}} = 0.1$ in our experiments. We find that the behaviour cloning loss is essential for the Assistive Itch Scratching task. Under this hyperparameter setting, we train for 200 iterations. During each iteration, we collect 19,200 state-action transitions, which is evenly divided into 20 mini-batches. Each mini-batch is fed to the base policy and encoder for 30 rounds to compute the loss and error for back-propagation. We set the learning rate to be $0.00005$. 

For PALM prediction training, we use a three-layer decoder with hidden size 12 to predict the next human action from the hidden state from the encoder. To implement the KL regularization, follow the standard VAE approach. We use two linear networks to transform the encoder hidden state into $\mu$ and $\sigma$, which denote the mean and the standard deviation of the latent space. We then compute approximate KL divergence to normal distribution on this latent distribution.

{\noindent \textbf{PALM Test Time Optimization} At test time, we roll out the trained robot policy and collect data with the same user for 25 iterations, or 2,500 time steps. This amounts to 150 seconds of wall clock time. We then optimize for the prediction loss (including KL regularization term) using learning rate of 0.0001 for one to five steps, and use the one with the lowest loss. We empirically find the hyperparameters by doing the same process with humans from the training distribution, where we collect a mini training set and mini evaluation set both of 25 iterations. We use the mini training set to find the learning rate and use the evaluation set to ensure there is no over-fitting.

\noindent \textbf{RILI/LILI Training} We follow a similar approach to PALM, except that we learn to predict the next state $o_{t+1}$ and scalar reward.

\noindent \textbf{RMA Training} We follow the two-phase training procedure in \cite{kumar2021rma}. Note that we find it crucial in phase 2 to train the encoder with on-policy data, meaning that the regression data is collected by rolling out actions output by the ``recurrent learner'', not the trained network from phase 1. The phase 1 network is used simply for generating labels.

\noindent \textbf{RNN/MLP Training}. For RNN, we directly feed the hidden state of the recurrent encoder to the policy network. The architectural difference between RNN and PALM is that we do not concatenate the current observation $o_t$ to the encoded output. To ensure that the policy has at least the same capacity as PALM, we use a base policy with the same number of parameters as in PALM.

\section{Ablation Studies}
\label{appendix_sec:ablation}

\begin{table}[h]
  \label{table:ablation-study}
  \centering
    \begin{tabular}{ l  l  l l l  }
     \hline
     \algabbr Baselines & Reach & Itch $\dist^1$ & Itch $\dist^2$ & Itch $\dist^3$\\
     \hline
      \algabbr test optim &   \textbf{0.43 $\pm$ 0.13}  & \textbf{0.51 $\pm$ 0.08} & \textbf{0.50 $\pm$ 0.02} & \textbf{0.29 $\pm$ 0.02} \\
      \algabbr w/o test optim & 0.38 $\pm$ 0.10 & 0.48 $\pm$ 0.08 & 0.49 $\pm$ 0.02 & 0.26 $\pm$ 0.01\\
      No $\Loss_{\textnormal{pred}}$  & 0.30 $\pm$ 0.05 & 0.50 $\pm$ 0.08 & 0.45 $\pm$ 0.02 & 0.25 $\pm$ 0.01\\
      $c_\textnormal{KL} = 0$         & 0.32 $\pm$ 0.08 & 0.47 $\pm$ 0.04 & 0.46 $\pm$ 0.01 & 0.23 $\pm$ 0.02\\
      Frozen $\encoder$  & 0.24 $\pm$ 0.04 & 0.21 $\pm$ 0.06 & 0.15 $\pm$ 0.07 & 0.11 $\pm$ 0.05 \\
     \hline
    \end{tabular}
    \caption{Normalized Reward on $\dtest$, standard deviation over 3 seeds.}
\end{table}

We include ablation studies of PALM in the main experiment in \sref{sec:experiments}, where we study the effect of test-time optimization, prediction loss, KL regularization and jointly training encoder $\encoder$ and policy $\pi$.

In the assistive reaching experiment, we observe that test-time optimization, $\Loss_{\textnormal{pred}}$, KL regularization, and joint training all contribute to the OOD performance.

In the assistive itch scratching experiment, test time optimization and $\Loss_{\textnormal{pred}}$ improve experiment results in all the settings. Applying KL regularization provides some gain in the complex distribution $\dist_3$, but does not lead to improvement in simpler distribution $\dist_1$ and $\dist_2$.

\section{Additional Experiment: Bed Bathing Task}
\label{appendix_sec:additional_bed}
We further evaluate the performance of PALM in another assistive robotics task: robot-assisted bathing. This task is a modified version of the bathing task introduced in Assistive Gym \cite{erickson2020assistive}. In this task, we have a human lying on a tilted bed, with a robot mounted on the nightstand. The human can move their right arm, and there is an identified patch of skin to be cleaned. Unlike the original bed bathing task, where the entire right arm is covered in target points to be cleaned, we instead initialize a fixed size region of points to be cleaned at a uniform randomly selected location along the surface of the right forearm. The patch spans 10 centimeter along and 150 degrees around the forearm. Only the human knows the center position of the patch, and the robot must infer the location of the patch based on observations of the human motion. The points along the body are cleaned whenever the robot initiates contact with the spot using its end-effector and applies a positive normal force. The task reward is based on how many points are cleaned.

\begin{figure}[h]
  \centering
    \includegraphics[width=\linewidth]{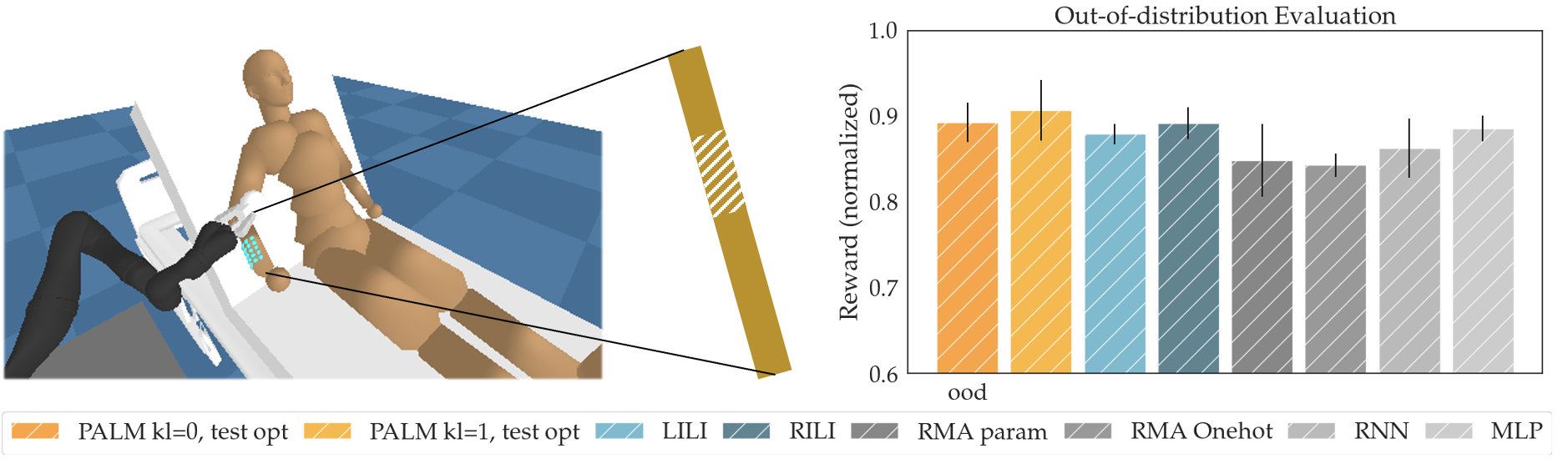}
    \small \caption{Visualization of the bed bathing task (left), where the human lies on a tilted bed with a table-mounted robot to clean an area on their arm. The area is random initialized (middle) on the forearm, where we hold out a quarter of the length as the held out distribution. The results of the held-out distribution is visualized on the right.}
    \label{fig:assistive_itch_humans}
\end{figure}

We generate synthetic humans by co-optimizing humans and robots conditioned on the centroid position of the region to be cleaned. During co-optimization, we adopt the formulation in \cite{erickson2020assistive} where both the human and the robot observe the centroid position along the human forearm. This helps induce collaborative behaviour for the human and robot policies, where we then use the resulting human as the synthetic population. During training, we blind of robot policy of the centroid position. We use the co-optimized robot (observes the centroid position) as the oracle for querying expert actions for Behaviour Cloning. We sample 18 humans from the training distribution, and save 6 humans from the held-out distribution as the out-of-distribution evaluation. We use the same hyperparameters as the itch scratching experiment.

As we see in the results in \figref{fig:assistive_itch_humans}, PALM achieves better out-of-distribution results compared to the baseline methods. We also observe that this performance gap is smaller than the itch scratching task. We believe this is due to the nature of the bed bathing task, where a robot controller that maintains contact with the human's forearm can be sufficient for solving the task if the human policy learns to move and rotate their forearm accordingly to help the robot.

\end{document}